\documentclass[]{article}
\usepackage{lmodern}
\usepackage{amssymb,amsmath}
\usepackage{ifxetex,ifluatex}
\usepackage{fixltx2e} 
\ifnum 0\ifxetex 1\fi\ifluatex 1\fi=0 
  \usepackage[T1]{fontenc}
  \usepackage[utf8]{inputenc}
\else 
  \ifxetex
    \usepackage{mathspec}
  \else
    \usepackage{fontspec}
  \fi
  \defaultfontfeatures{Ligatures=TeX,Scale=MatchLowercase}
\fi
\IfFileExists{upquote.sty}{\usepackage{upquote}}{}
\IfFileExists{microtype.sty}{%
\usepackage{microtype}
\UseMicrotypeSet[protrusion]{basicmath} 
}{}
\usepackage{hyperref}
\PassOptionsToPackage{usenames,dvipsnames}{color} 
\hypersetup{unicode=true,
            pdftitle={MDP environments for the OpenAI Gym},
            pdfauthor={Andreas Kirsch blackhc@gmail.com},
            colorlinks=true,
            linkcolor=Maroon,
            citecolor=Blue,
            urlcolor=Blue,
            breaklinks=true}
\usepackage{color}
\usepackage{fancyvrb}

\DefineVerbatimEnvironment{Highlighting}{Verbatim}{commandchars=\\\{\}}
\newenvironment{Shaded}{}{}

\newcommand{\DecValTok}[1]{\textcolor[rgb]{0.25,0.63,0.44}{{#1}}}

\newcommand{\StringTok}[1]{\textcolor[rgb]{0.25,0.44,0.63}{{#1}}}

\newcommand{\ImportTok}[1]{{#1}}

\newcommand{\VariableTok}[1]{\textcolor[rgb]{0.10,0.09,0.49}{{#1}}}
\newcommand{\ControlFlowTok}[1]{\textcolor[rgb]{0.00,0.44,0.13}{\textbf{{#1}}}}
\newcommand{\OperatorTok}[1]{\textcolor[rgb]{0.40,0.40,0.40}{{#1}}}
\newcommand{\BuiltInTok}[1]{{#1}}

\newcommand{\NormalTok}[1]{{#1}}
\usepackage{graphicx,grffile}
\makeatletter
\def\maxwidth{\ifdim\Gin@nat@width>\linewidth\linewidth\else\Gin@nat@width\fi}
\def\maxheight{\ifdim\Gin@nat@height>\textheight\textheight\else\Gin@nat@height\fi}
\makeatother
\setkeys{Gin}{width=\maxwidth,height=\maxheight,keepaspectratio}
\IfFileExists{parskip.sty}{%
\usepackage{parskip}
}{
\setlength{\parindent}{0pt}
\setlength{\parskip}{6pt plus 2pt minus 1pt}
}
\providecommand{\tightlist}{%
  \setlength{\itemsep}{0pt}\setlength{\parskip}{0pt}}
\ifx\paragraph\undefined\else
\let\oldparagraph\paragraph
\renewcommand{\paragraph}[1]{\oldparagraph{#1}\mbox{}}
\fi
\ifx\subparagraph\undefined\else
\let\oldsubparagraph\subparagraph
\renewcommand{\subparagraph}[1]{\oldsubparagraph{#1}\mbox{}}
\fi

\title{MDP environments for the OpenAI Gym}
\author{Andreas Kirsch
\href{mailto:blackhc@gmail.com}{\nolinkurl{blackhc@gmail.com}}}
\date{}

\begin{document}
\maketitle
\begin{abstract}
The OpenAI Gym provides researchers and enthusiasts with simple to use
environments for reinforcement learning. Even the simplest environment
have a level of complexity that can obfuscate the inner workings of RL
approaches and make debugging difficult. This whitepaper describes a
Python framework\footnote{\url{https://github.com/BlackHC/mdp}} that
makes it very easy to create simple Markov-Decision-Process environments
programmatically by specifying state transitions and rewards of
deterministic and non-deterministic MDPs in a domain-specific language
in Python. It then presents results and visualizations created with this
MDP framework.
\end{abstract}

\section{Introduction}\label{introduction}

In reinforcement learning{[}\protect\hyperlink{ref-Sutton1998}{6}{]},
agents learn to maximize accumulated rewards from an environment that
they can interact with by observing and taking actions. Usually, these
environments satisfy a Markov property and are treated as \emph{Markov
Decision Processes}
(\emph{MDPs}){[}\protect\hyperlink{ref-Puterman2005}{4}{]}.

The OpenAI
Gym{[}\protect\hyperlink{ref-DBLP:journalsux2fcorrux2fBrockmanCPSSTZ16}{1}{]}
is a standardized and open framework that provides many different
environments to train agents against through a simple API.

Even the simplest of these environments already has a level of
complexity that is interesting for research but can make it hard to
track down bugs. However, the gym provides four very simple environments
that are useful for testing. The \texttt{gym.envs.debugging} package
contains a one-round environment with deterministic rewards and one with
non-deterministic rewards, and a two-round environment with
deterministic rewards and another one with non-deterministic rewards.
The author has found these environments very useful for smoke-testing
code changes.

This whitepaper introduces a Python framework that makes it very easy to
specify simple MDPs like the ones described above in an extensible way.
With it, one can validate that agents converge correctly as well as
examine other properties.

\section{MDP framework}\label{mdp-framework}

\subsection{Specification of MDPs}\label{specification-of-mdps}

MDPs are Markov processes that are augmented with a reward function and
discount factor. An MDP can be fully specified by a tuple of:

\begin{itemize}
\tightlist
\item
  a finite set of states,
\item
  a finite set of actions,
\item
  a matrix that specifies probabilities of transitions to a new state
  for a given a state and action,
\item
  a reward function that specifies the reward for a given action taken
  in a certain state, and
\item
  a discount rate.
\end{itemize}

The reward function can be either deterministic, or it can be a
probability distribution.

Within the framework, MDPs can be specified in Python using a simple
\emph{domain-specific language} (\emph{DSL}). For example, the one-round
deterministic environment defined in
\texttt{gym.envs.debugging.one\_round\_deterministic\_reward} could be
specified as follows:

\begin{Shaded}
\begin{Highlighting}[]
\ImportTok{from} \NormalTok{blackhc.mdp }\ImportTok{import} \NormalTok{dsl}

\NormalTok{start }\OperatorTok{=} \NormalTok{dsl.state()}
\NormalTok{end }\OperatorTok{=} \NormalTok{dsl.terminal_state()}

\NormalTok{action_0 }\OperatorTok{=} \NormalTok{dsl.action()}
\NormalTok{action_1 }\OperatorTok{=} \NormalTok{dsl.action()}

\NormalTok{start }\OperatorTok{&} \NormalTok{(action_0 }\OperatorTok{|} \NormalTok{action_1) }\OperatorTok{>} \NormalTok{end}
\NormalTok{start }\OperatorTok{&} \NormalTok{action_1 }\OperatorTok{>} \NormalTok{dsl.reward(}\DecValTok{1}\NormalTok{.)}
\end{Highlighting}
\end{Shaded}

The DSL is based on the following grammar (using
EBNF{[}\protect\hyperlink{ref-ebnf}{5}{]}):

\begin{verbatim}
TRANSITION ::= STATE '&' ACTION '>' OUTCOME
OUTCOME ::= (REWARD | STATE) ['*' WEIGHT]
ALTERNATIVES ::= ALTERNATIVE ('|' ALTERNATIVE)* 
\end{verbatim}

For a given state and action, outcomes can be specified. Outcomes are
state transitions or rewards. If multiple state transitions or rewards
are specified for the same state and action, the MDP is
non-deterministic and the state transition (or reward) are determined
using a categorical distribution. By default, each outcome is weighted
uniformly, except if specified otherwise by either having duplicate
transitions or by using an explicit weight factor.

For example, to specify that a state receives a reward of +1 or -1 with
equal probability and does not change states with probability \(3/4\)
and only transitions to the next state with probability \(1/4\), we
could write:

\begin{Shaded}
\begin{Highlighting}[]
\NormalTok{state }\OperatorTok{&} \NormalTok{action }\OperatorTok{>} \NormalTok{dsl.reward(}\OperatorTok{-}\DecValTok{1}\NormalTok{.) }\OperatorTok{|} \NormalTok{dsl.reward(}\DecValTok{1}\NormalTok{.)}
\NormalTok{state }\OperatorTok{&} \NormalTok{action }\OperatorTok{>} \NormalTok{state }\OperatorTok{*} \DecValTok{3} \OperatorTok{|} \NormalTok{next_state}
\end{Highlighting}
\end{Shaded}

Alternatives are distributive with respect to both conjunctions
(\texttt{\&}) and outcome mappings (\texttt{\textgreater{}}), so:

\begin{verbatim}
(a | b) & (c | d) > (e | f) ===
(a & c > e) | (a & c > f) | (a & d > e) | 
(a & d > f) | (b & c > e) | ... 
\end{verbatim}

Alternatives can consist of states, actions, outcomes, conjunctions or
partial transitions. For example, the following are valid alternatives:

\begin{verbatim}
stateA & actionA | stateB & actionB
(actionA > stateC) | (actionB > stateD)
\end{verbatim}

As the DSL is implemented within Python, operator overloading is used to
implement the semantics. Operator precedence is favorable as \texttt{*}
has higher precedence than \texttt{\&}, which has higher precedence than
\texttt{\textbar{}}, which has higher precedence than
\texttt{\textgreater{}}. This allows for a natural formulation of
transitions.

\subsection{Conventional API}\label{conventional-api}

The framework also supports specifying an MDP using a conventional API
as DSLs are not always preferred.

\begin{Shaded}
\begin{Highlighting}[]
\ImportTok{from} \NormalTok{blackhc }\ImportTok{import} \NormalTok{mdp}

\NormalTok{spec }\OperatorTok{=} \NormalTok{mdp.MDPSpec()}
\NormalTok{start }\OperatorTok{=} \NormalTok{spec.state(}\StringTok{'start'}\NormalTok{)}
\NormalTok{end }\OperatorTok{=} \NormalTok{spec.state(}\StringTok{'end'}\NormalTok{, terminal_state}\OperatorTok{=}\VariableTok{True}\NormalTok{)}
\NormalTok{action_0 }\OperatorTok{=} \NormalTok{spec.action()}
\NormalTok{action_1 }\OperatorTok{=} \NormalTok{spec.action()}

\NormalTok{spec.transition(start, action_0, mdp.NextState(end))}
\NormalTok{spec.transition(start, action_1, mdp.NextState(end))}
\NormalTok{spec.transition(start, action_1, mdp.Reward(}\DecValTok{1}\NormalTok{))}
\end{Highlighting}
\end{Shaded}

\subsection{Visualization}\label{visualization}

To make debugging easier, MDPs can be converted to \texttt{networkx}
graphs{[}\protect\hyperlink{ref-networkx}{3}{]} and rendered using
\texttt{pydotplus} and
\texttt{GraphViz}{[}\protect\hyperlink{ref-graphviz}{2}{]}.

\begin{Shaded}
\begin{Highlighting}[]
\ImportTok{from} \NormalTok{blackhc }\ImportTok{import} \NormalTok{mdp}
\ImportTok{from} \NormalTok{blackhc.mdp }\ImportTok{import} \NormalTok{example}

\NormalTok{spec }\OperatorTok{=} \NormalTok{example.ONE_ROUND_DMDP}

\NormalTok{spec_graph }\OperatorTok{=} \NormalTok{spec.to_graph()}
\NormalTok{spec_png }\OperatorTok{=} \NormalTok{mdp.graph_to_png(spec_graph)}

\NormalTok{mdp.display_mdp(spec)}
\end{Highlighting}
\end{Shaded}

\begin{figure}[htbp]
\centering
\includegraphics[width=1.00000in]{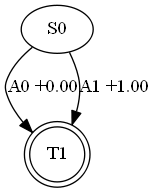}
\caption{One round deterministic MDP}
\end{figure}

\subsection{Optimal values}\label{optimal-values}

The framework also contains a small module that can compute the optimal
value functions using linear programming.

\begin{Shaded}
\begin{Highlighting}[]
\ImportTok{from} \NormalTok{blackhc.mdp }\ImportTok{import} \NormalTok{lp}
\ImportTok{from} \NormalTok{blackhc.mdp }\ImportTok{import} \NormalTok{example}

\NormalTok{solver }\OperatorTok{=} \NormalTok{lp.LinearProgramming(example.ONE_ROUND_DMDP)}
\BuiltInTok{print}\NormalTok{(solver.compute_q_table())}
\BuiltInTok{print}\NormalTok{(solver.compute_v_vector())}
\end{Highlighting}
\end{Shaded}

\subsection{Gym environment}\label{gym-environment}

An environment that is compatible with the OpenAI Gym can be created
easily by using the \texttt{to\_env()} method. It supports rendering
into Jupyter notebooks, as RGB array for storing videos, and as png byte
data.

\begin{Shaded}
\begin{Highlighting}[]
\ImportTok{from} \NormalTok{blackhc }\ImportTok{import} \NormalTok{mdp}
\ImportTok{from} \NormalTok{blackhc.mdp }\ImportTok{import} \NormalTok{example}

\NormalTok{env }\OperatorTok{=} \NormalTok{example.MULTI_ROUND_NMDP.to_env()}

\NormalTok{env.reset()}
\NormalTok{env.render()}

\NormalTok{is_done }\OperatorTok{=} \VariableTok{False}
\ControlFlowTok{while} \OperatorTok{not} \NormalTok{is_done:}
    \NormalTok{state, reward, is_done, _ }\OperatorTok{=} \NormalTok{env.step(env.action_space.sample())}
    \NormalTok{env.render()}
\end{Highlighting}
\end{Shaded}

\begin{figure}[htbp]
\centering
\includegraphics[width=2.00000in]{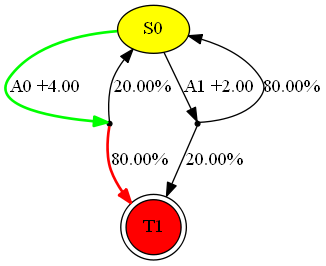}
\caption{env.render() of \texttt{example.MULTI\_ROUND\_NMDP}
\label{multi_ndmp}}
\end{figure}

\subsection{Examples}\label{examples}

The \texttt{blackhc.mdp.example} package provides 5 MDPs. Four of them
match the ones in \texttt{gym.envs.debugging}, and the fifth one is
depicted in figure \ref{multi_ndmp}.

\section{Contribution}\label{contribution}

A framework to specify MDPs using a domain-specific language in Python
was presented. The MDPs can be visualized in Jupyter notebooks and are
compatible with the OpenAI Gym.

\section{Acknowledgements}\label{acknowledgements}

Thanks to OpenAI for developing and providing the OpenAI Gym, and thanks
to John Maguire for feedback.

\section*{References}\label{references}
\addcontentsline{toc}{section}{References}

\hypertarget{refs}{}
\hypertarget{ref-DBLP:journalsux2fcorrux2fBrockmanCPSSTZ16}{}
{[}1{]} Brockman, G., Cheung, V., Pettersson, L., Schneider, J.,
Schulman, J., Tang, J. and Zaremba, W. 2016. OpenAI gym. \emph{CoRR}.
abs/1606.01540, (2016).

\hypertarget{ref-graphviz}{}
{[}2{]} Gansner, E.R. and North, S.C. 2000. An open graph visualization
system and its applications to software engineering. \emph{SOFTWARE -
PRACTICE AND EXPERIENCE}. 30, 11 (2000), 1203--1233.

\hypertarget{ref-networkx}{}
{[}3{]} Hagberg, A.A., Schult, D.A. and Swart, P.J. 2008. Exploring
network structure, dynamics, and function using NetworkX.
\emph{Proceedings of the 7th python in science conference (sciPy2008)}
(Pasadena, CA USA, Aug. 2008), 11--15.

\hypertarget{ref-Puterman2005}{}
{[}4{]} Puterman, M.L. 2005. \emph{Markov decision processes : discrete
stochastic dynamic programming}. Wiley-Interscience.

\hypertarget{ref-ebnf}{}
{[}5{]} Scowen, R.S. 1998. \emph{Extended bNF-a generic base standard}.
Technical report, ISO/IEC 14977.
http://www.cl.cam.ac.uk/mgk25/iso-14977.pdf.

\hypertarget{ref-Sutton1998}{}
{[}6{]} Sutton, R.S. and Barto, A.G. 1998. \emph{Reinforcement learning
: an introduction}. MIT Press.

\end{document}